\documentclass{llncs}
\usepackage{amssymb}
\usepackage{color, soul}

\usepackage{graphicx}
\usepackage{subfig}
\usepackage{amsmath}
\usepackage{mathtools}
\usepackage{gensymb}
\usepackage{lipsum,wrapfig}
\usepackage{bm}
\usepackage{bbold}
\usepackage{paralist}
\usepackage{tabularx}
\usepackage{cite}

\newcommand{\x}{{\mathbf{x}}}
\newcommand{\xsupi}{{\mathbf{x}^i}}

\newcommand{\y}{{\mathbf{y}}}
\newcommand{\ysz}{{\mathbf{y}_0}}
\newcommand{\yszsupi}{{\mathbf{y}_0^i}}
\newcommand{\ysone}{{\mathbf{y}_1}}

\newcommand{\ysT}{{\mathbf{y}_T}}
\newcommand{\ysTsupi}{{\mathbf{y}^i_{T_i}}}

\newcommand{\yst}{{\mathbf{y}_t}}

\newcommand{\ystmo}{{\mathbf{y}_{t-1}}}
\newcommand{\ystmostar}{{\mathbf{y}_{t-1}^{*}}}

\newcommand{\ysi}{{\mathbf{y}_i}}
\newcommand{\ysimo}{{\mathbf{y}_{i-1}}}

\newcommand{\ssz}{{s_0}}
\newcommand{\sszsupi}{{s_0^i}}
\newcommand{\ssone}{{s_1}}

\newcommand{\ssT}{{s_T}}
\newcommand{\ssTsupi}{{s^i_{T_{i}}}}

\newcommand{\ssTimosupi}{{s^i_{T_{i-1}}}}

\newcommand{\sst}{{s_t}}
\newcommand{\sststar}{{s_t^{*}}}
\newcommand{\sstmo}{{s_{t-1}}}
\newcommand{\sstmostar}{{s^{*}_{t-1}}}
\newcommand{\ssi}{{s_i}}

\newcommand{\hst}{{\mathbf{h}_t}}

\newcommand{\params}{{\bm{\theta}}}

\newcommand{\paramsh}{{\bm{\theta}_h}}
\newcommand{\paramsy}{{\bm{\theta}_y}}
\newcommand{\paramss}{{\bm{\theta}_s}}
\newcommand{\paramsstar}{{\bm{\theta}^{*}}}

\newcommand{\argmin}{\mathop{\mathrm{argmin}}}
\newcommand{\argmax}{\mathop{\mathrm{argmax}}}

\newcommand{\indicator}{{\mathbb{1}}}

\def\etal{\emph{et al. }}

\usepackage[mynonotes]{optional}

\begin{document}

% spacing before and after equations
%\abovedisplayskip=3pt
%\belowdisplayskip=3pt
%
\pagestyle{headings}  % switches on printing of running heads
\title{Iterative Segmentation from Limited Training Data: Applications to Congenital Heart Disease}
\titlerunning{Iterative Segmentation from Limited Training Data: Applications to CHD}  % abbreviated title (for running head)
\author{Authors}

\author{Danielle F. Pace\inst{1} \and Adrian V. Dalca\inst{1,2,3} \and Tom Brosch\inst{4} \and Tal Geva\inst{5,6} \and Andrew J. Powell\inst{5,6} \and J\"{u}rgen Weese\inst{4} \and Mehdi H. Moghari\inst{5,6} \and Polina Golland\inst{1}}

\index{Pace, Danielle F.}
\index{Dalca, Adrian V.} 
\index{Brosch, Tom}
\index{Geva, Tal} 
\index{Powell, Andrew J.} 
\index{Weese, J\"{u}rgen}
\index{Moghari, Mehdi H.} 
\index{Golland, Polina} 

\authorrunning{D.F. Pace \etal} % abbreviated author list (for running head)

\institute{Computer Science and Artificial Intelligence Lab, MIT (\email{dfpace@mit.edu})\\
\and Martinos Center for Biomedical Imaging, Massachusetts General Hospital, HMS\\
\and School of Electrical and Computer Engineering, Cornell University\\
\and Philips Research Laboratories\\
\and Department of Cardiology, Boston Children's Hospital\\
\and Department of Pediatrics, Harvard Medical School}

\maketitle              % typeset the title of the contribution

\begin{abstract}
We propose a new iterative segmentation model which can be accurately learned from a small dataset. A common approach is to train a model to directly segment an image, requiring a large collection of manually annotated images to capture the anatomical variability in a cohort. In contrast, we develop a segmentation model that recursively evolves a segmentation in several steps, and implement it as a recurrent neural network. We learn model parameters by optimizing the intermediate steps of the evolution in addition to the final segmentation. To this end, we train our segmentation propagation model by presenting incomplete and/or inaccurate input segmentations paired with a recommended next step. Our work aims to alleviate challenges in segmenting heart structures from cardiac MRI for patients with congenital heart disease (CHD), which encompasses a range of morphological deformations and topological changes. We demonstrate the advantages of this approach on a dataset of 20 images from CHD patients, learning a model that accurately segments individual heart chambers and great vessels. Compared to direct segmentation, the iterative method yields more accurate segmentation for patients with the most severe CHD malformations.
\end{abstract}

\section{Introduction}

We aim to provide whole heart segmentation in cardiac MRI for patients with congenital heart disease (CHD). This involves delineating the heart chambers and great vessels \cite{zhuang_challenges_2013}, and promises to enable patient-specific heart models for surgical planning in CHD \cite{pace_interactive_2015}. CHD encompasses a vast range of cardiac malformations and topological changes. Defects can include holes in the heart walls (septal defects), great vessels connected to the wrong chamber (e.g., double outlet right ventricle; DORV), dextrocardia (left-right flip), duplication of a great vessel, a single ventricle, and/or prior surgeries creating additional atypical connections. In MRI, different chambers and great vessels locally appear very similar to each other, and there is little or no contrast at the valves and thin walls separating neighboring structures. Finally, labeled training data is very limited. This precludes modeling each CHD subtype separately in an attempt to reduce variability. Moreover, patients with unique combinations of defects and prior surgeries defy categorization. Beyond our application, limited training data is to be expected for new applications of medical imaging not yet in widespread clinical practice. This necessitates development of methods that generalize well from small, imbalanced datasets, possibly also incorporating user interaction.

State-of-the-art methods use a convolutional neural network (CNN) to directly outline all chambers and vessels in one step \cite{payer_multi-label_2017,wang_automatic_2017}. However, CNNs for CHD have largely been limited to segmenting the blood pool and myocardium \cite{wolterink_dilated_2016,yu_3d_2016}. Direct co-segmentation of all major cardiac structures works well when applied to adult-onset heart disease, which induces much less severe shape changes compared to CHD. However, it fails completely on held-out subjects with severe CHD malformations after training with our small dataset of CHD patients.

We develop an iterative segmentation approach that evolves a segmentation over several steps in a prescribed way and automatically estimates when to stop, beginning from a single seed for each structure placed by the user. An iterative method can operate more locally, better maintain each structure's connectivity, and propagate information from distant landmarks, similar to traditional snakes, level sets and particle filters \cite{Sonka-et-al-2008}. We employ a recurrent neural network (RNN) \cite{Goodfellow-et-al-2016}, which uses context to grow the segmentation appropriately even in areas of low contrast. Deep learning research has indeed focused on segmenting a single image iteratively. Examples include recursive refinement of the entire segmentation map \cite{Pinheiro_recurrent_2014,zhou2017fixed}, sequential completion of different instances, regions or fields of view \cite{Ren_2017_CVPR,Banica_2015_CVPR,januszewski_high-precision_2018}, slice-by-slice analysis \cite{zheng_3d_2018} and networks modeling level set evolution \cite{Chakravarty_race_2018}. These methods condition on a previous partial solution to make progress towards the final output. This simplified task may enable training from smaller datasets.

We train the model by minimizing a loss over a training dataset of example segmentation \emph{trajectories}. Maximizing the likelihood of observed sequences is known as teacher forcing \cite{Goodfellow-et-al-2016,williams_learning_1989}. For example, we may require vessel segmentation to proceed at a constant rate along the vessel centerline, or a heart chamber segmentation to dilate outwards. Even if the stopping prediction is incorrect, since the segmentation evolution follows a prescribed pattern it is likely that one of the intermediate segmentations will be accurate. In contrast, using the final segmentation alone could lead to unpredictable growth patterns. Teacher forcing also leads to a simplified optimization over decoupled time steps, avoiding back-propagation through time.

We focus on segmenting the aorta (a representative great vessel) and the left ventricle (a representative cardiac chamber). We validate our iterative segmentation approach using a dataset of 20 CHD patients, and compare it to direct segmentation methods which we have developed for this problem.

%\begingroup
%\setlength\intextsep{0pt}
%\begin{wrapfigure}[18]{l}{1.5in}
%\centering
%\includegraphics[width=1.5in,height=3in]{todo.jpg}
%\caption{Direct segmentation fails for severely malformed hearts.}
%\label{fig:badCases}
%\end{wrapfigure}

\section{Iterative Segmentation Model}

Given an input image $\x$ defined on the domain $\Omega$, we seek a segmentation label map $\y$ that assigns one of $L$ anatomical labels to each voxel in $\x$. 

\textbf{Generative model:} We model the segmentation $\y$ as the endpoint of a sequence of segmentations $\ysz, \ldots, \ysT$, where $\yst : \Omega \rightarrow \{1, \ldots, L\}$ for time steps $t=0, \ldots, T$. The intermediate segmentations $\yst$ capture a growing part of the anatomy of interest. In practice, the initial segmentation map $\ysz$ is created by centering a small sphere around an initial seed point placed by the user. 

The number of iterations required to achieve an accurate segmentation depends on the shape and size of the object being segmented. To capture this, we introduce a sequence of indicator variables $\ssz, \ldots, \ssT$, where $\sst \in \{0, 1\}$ specifies whether the segmentation is completed at time step $t$. If $\sst = 1$, then $\yst$ is deemed the final segmentation and we set $\ysi = \ysimo$ and $\ssi = 1$ for all $i > t$.

Given an image and an initial segmentation, the inference task is to compute $p(\ysT, \ssT \vert \x, \ysz, \ssz = 0)$. We assume that the segmentations $\{ \yst \}$ and stopping indicators $\{ \sst\}$ follow a first order Markov chain given the input image:
\begin{equation}
p(\yst, \sst \vert \x, \ysz, \ldots, \ystmo, \ssz, \ldots, \sstmo) = p(\yst, \sst \vert \x, \ystmo, \sstmo),
\label{eq:markov}
\end{equation}
\begin{equation}
p(\yst, \sst \vert \x, \ysz, \ssz) = \sum_{\ystmo} \sum_{\sstmo} p(\yst, \sst \vert \x, \ystmo, \sstmo) \cdot p(\ystmo, \sstmo \vert \x, \ysz, \ssz).
\label{eq:marg}
\end{equation}

\vspace{3pt}
\textbf{Transition probability model:} We must define the transition probability $p(\yst, \sst \vert \x, \ystmo, \sstmo)$ to complete the recursion in eqn. (\ref{eq:marg}). There are two possible cases: $\sstmo = 1$ and $\sstmo = 0$. Based on the definition of $\sstmo$, we obtain
\begin{equation}
\begin{split}
p(\yst, \sst \vert \x, \ystmo, \sstmo = 1) & = \indicator(\yst = \ystmo) \cdot \indicator(\sst = 1),
\end{split}
\end{equation}
where $\indicator(\cdot)$ denotes the indicator function. To compute $p(\yst, \sst \vert \x, \ystmo, \sstmo = 0)$, we introduce a latent representation $\hst = h(\x, \ystmo)$ that jointly captures all of the necessary information from image $\x$ and previous segmentation $\ystmo$. Intuitively, predicting whether the segmentation $\yst$ is complete given $\x$ can be performed by examining whether $\ystmo$ is ``almost'' complete. Therefore, the segmentation $\yst$ and stopping indicator $\sst$ are conditionally independent given $\hst$:
\begin{equation}
p(\yst, \sst \vert \x, \ystmo, \sstmo = 0)= p(\yst, \sst \vert \hst) = p(\yst \vert \hst) \cdot p(\sst \vert \hst).
\label{eq:condsstmozero}
\end{equation}
We model the function $h(\x, \ystmo)$ and distributions $p(\yst \vert \hst)$ and $p(\sst \vert \hst)$ as stationary; they do not depend on the time step $t$.

\vspace{3pt}
\textbf{Learning:} We learn a representation of $p(\yst, \sst \vert \x, \ystmo, \sstmo = 0)$ given a training dataset of example desired trajectories of segmentations. Specifically, we consider a training dataset $\mathcal{D}$ of $N$ images $\{\xsupi\}_{i=1}^N$, each of which has a corresponding sequence of segmentations $\yszsupi, \ldots, \ysTsupi$ and of stopping indicators $\sszsupi, \ldots, \ssTsupi$, where $\sszsupi = \ldots = \ssTimosupi = 0$ and $\ssTsupi = 1$. The parameter values to be determined are $\params = \{\paramsh, \paramsy, \paramss\}$ corresponding to $h(\x, \ystmo ; \paramsh)$, $p(\yst \vert \hst ; \paramsy)$, and $p(\sst \vert \hst ; \paramss)$, respectively. We seek the parameter values that minimize the expected negative log-likelihood of the output segmentation and stopping indicator sequences given the image and initial conditions, i.e., $\paramsstar = \argmin_\params \mathcal{L}(\params)$,
\begin{equation}
\begin{split}
\mathcal{L}(\params) & = \mathbb{E}_{\x,  \ysz, \ldots, \ysT, \ssz, \ldots, \ssT \sim \mathcal{D}}\Big[- \log p(\ysone, \ldots, \ysT, \ssone, \ldots, \ssT \vert \x, \ysz, \ssz ; \params) \Big] \\
& = -\mathbb{E}\Big[ \sum_{t=1}^{T} \log p(\yst \vert h(\x, \ystmo ; \paramsh) ; \paramsy) + \log p(\sst \vert h(\x, \ystmo ; \paramsh) ; \paramss) \Big].
\label{eq:loss}
\end{split}
\end{equation}
Note that teacher forcing has lead to decoupled time steps. The first and second terms in the likelihood above penalize differences for the segmentations and the stopping indicators, respectively, between the predicted probabilities and the ground truth. In practice, we perform class rebalancing for both terms, and further supplement the segmentation loss by more strongly weighting pixels on the boundaries of the ground truth segmentation.

\vspace{3pt}
\setlength{\jot}{3pt}
\textbf{Inference:} Computing $p(\ysT, \ssT \vert \x, \ysz, \ssz = 0)$ via the recursion in eqn. (\ref{eq:marg}) is intractable due to the summation over all possible segmentations $\ystmo$. To approximate, we follow a widely accepted practice of using the most likely segmentation $\ystmostar$ and stopping indicator $\sstmostar$ as input to the subsequent computation:
\begin{equation}
\begin{split}
p(\yst, \sst \vert \x, \ysz, \ssz = 0 ; \params) & \approx p(\yst, \sst \vert \x, \ystmostar, \sstmostar ; \params),\\
\text{where } \ystmostar, \sstmostar & = \argmax_{\ystmo, \: \sstmo} p(\ystmo, \sstmo \vert \x, \ysz, \ssz = 0 ; \params).
\label{eq:inference}
\end{split}
\end{equation}
The segmentation is fully automatic given the initial seed. If the stopping indicator is predicted incorrectly, a user can manually override it by asking for more iterations or by choosing a segmentation from a previous step.

\begin{figure}[!t]
\captionsetup{belowskip=0pt}
\centering
\includegraphics[height=2.3in]{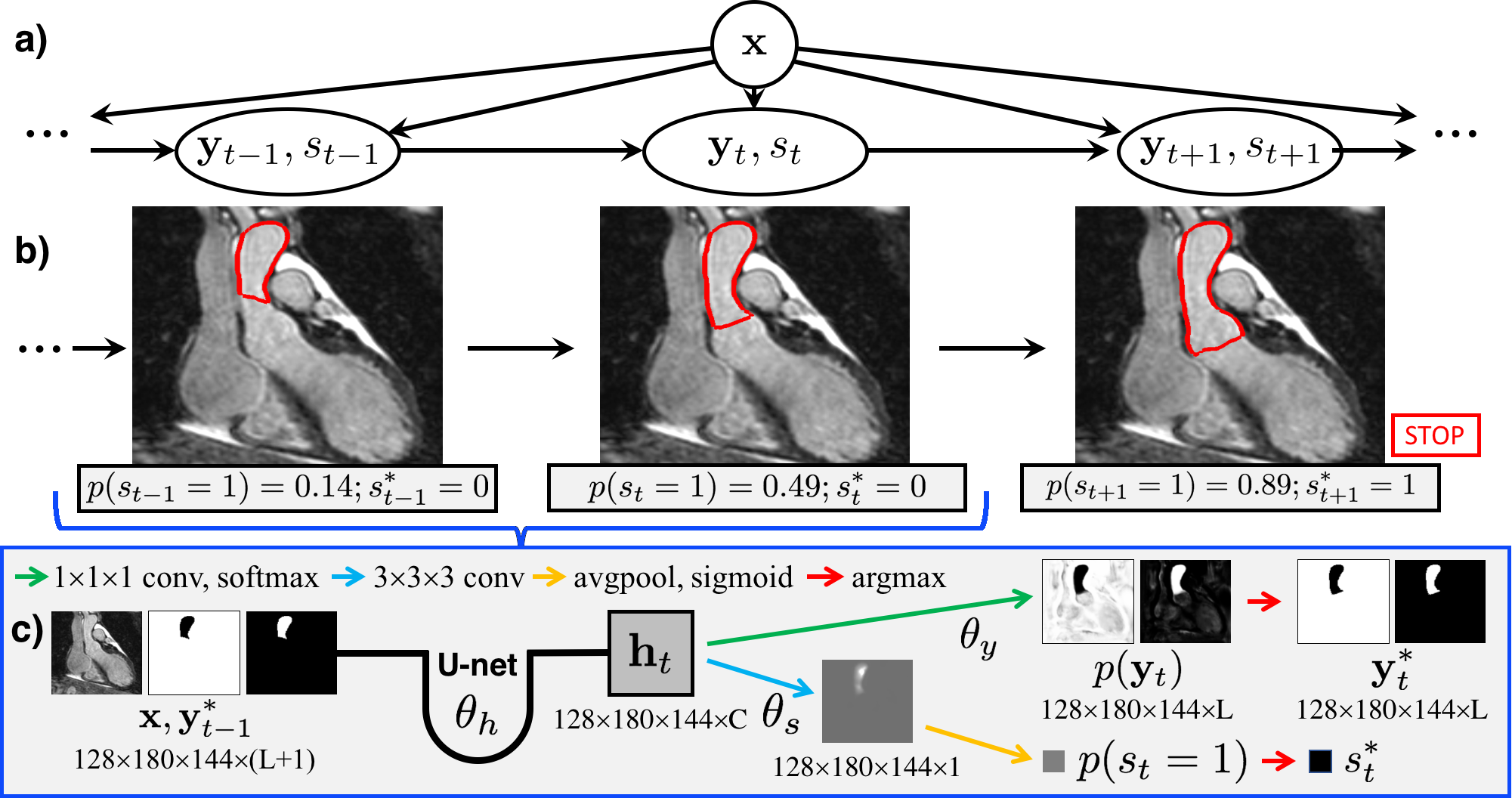} % was 2.5
\caption{Iterative segmentation as an RNN. (a) Generative model. (b) The RNN uses the same augmented U-net at each step to predict the next segmentation and stopping indicator. (c) Architecture details (conditioning dropped for clarity).}
\label{fig:schematic}
\end{figure}

\vspace{3pt}
\textbf{RNN:} We implement our iterative segmentation model as an RNN (Fig.~\ref{fig:schematic}), which is formed by connecting identical copies of an augmented 3D U-net \cite{ronneberger_u-net:_2015} trained to estimate $p(\yst, \sst \vert \x, \ystmo, \sstmo = 0)$. Thus, parameters are shared both spatially and temporally. At each step, the U-net inputs the image and the most likely segmentation from the previous step. This respects the Markov property in eqn.~(\ref{eq:markov}), unlike if any hidden layers were connected between successive steps. If the stopping indicator $\sststar = 1$, the segmentation propagation halts.

Our augmented U-net modeling $p(\yst, \sst \vert \x, \ystmo, \sstmo = 0)$ has $L+1$ input channels, containing the input image and a binary map for each of the $L$ labels in the segmentation $\ystmo$ (including the background). There are two outputs: the probability map for the segmentation $\yst$ (at each voxel, representing the parameters of the categorical distribution over $L$ labels), and the Bernoulli stopping parameter $p(\sst = 1 \vert \x, \ystmo, \sstmo = 0)$. Jointly predicting the segmentation and stopping indicator enables a smaller model compared to two separate networks.

The original U-net for image segmentation produces a final set of $C$ learned feature maps, which undergo $C \cdot L$ $1 \times 1 \times 1$ convolutions and a softmax activation to give the output segmentation probabilities. We use these $C$ learned feature maps as the latent joint representation $\hst = h(\x, \ystmo ; \paramsh)$. The U-net parameters can therefore be split into two sets. The parameters for the final $1 \times 1 \times 1$ convolutions are $\paramsy$ of $p(\yst \vert \hst ; \paramsy)$, and the remainder are $\paramsh$ of $h(\x, \ystmo ; \paramsh)$. The probability $p(\sst = 1 \vert \hst; \paramss)$ is computed by applying $C$ additional $3 \times 3 \times 3$ convolutions with parameters $\paramss$ to the feature maps in $\hst$, followed by a global average and sigmoid activation to yield a scalar in $\{0,1\}$.

\vspace{3pt}
\textbf{Generating segmentation trajectories:} Our training dataset of images and segmentation trajectories is derived from a collection of paired images and complete segmentations. Several acceptable trajectories exist for each pair, e.g., starting from different initial seeds. To this end, at the beginning of each epoch a random tuple $(\ystmo, \yst, \sst)$ is generated for each image. These tuples all follow the same principle that we want the network to learn.

As a concrete example, the trajectories used in our experiments are as follows. For the \emph{aorta}, the segmentation grows from the seed along the vessel centerline, by a random distance to form $\ystmo$ and an additional 10 pixels for $\yst$. The seed is placed in the descending aorta, and the endpoint is at the valve where the aorta connects to a left or right ventricle. This seed could be automatically detected in the future, and the lack of contrast at the valve provides a challenging test case for our automatic stopping. For the \emph{left ventricle}, we randomly place the seed in the center region of the chamber, and perform a random number of dilations to form $\ystmo$, and 3 more dilations to form $\yst$.

\vspace{3pt}
\textbf{Data Augmentation:} Data augmentation is essential to prevent overfitting on a small training dataset. We mimic the diversity of heart shapes and sizes, global intensity changes caused by inhomogeneity artifacts, and noise induced by elevated heart rates or arrhythmias. We apply random rigid and nonrigid transformations, random constant intensity shifts, and random additive Gaussian noise. We also investigate including random left-right (L-R) and anterior-posterior (A-P) flips, to better handle dextrocardia or other cardiac malpositions, since in these cases the left ventricle may lie on the right side of the body.

If the augmented U-net for $p(\yst, \sst \vert \x, \ystmo, \sstmo = 0)$ is trained solely using error-free segmentations $\ystmo$, then it may not operate well on its own imperfect intermediate results at test time. We increase robustness by performing additional data augmentation on the input segmentations $\ystmo$. We corrupt these segmentations by applying random nonrigid deformations, and by inserting random blob-like structures that vary in number, location and size and are attached to the segmentation foreground or free-floating. Since the target segmentation $\yst$ remains unchanged, the model learns to correct mistakes in its input.

\section{Experimental Validation}

We evaluate our iterative segmentation and tailored direct segmentation methods, focusing on segmenting the aorta and left ventricle (LV) of CHD patients.

\vspace{3pt}
\textbf{Data:}  We use the HVSMR dataset of 20 MRI scans from patients with a variety of congenital heart defects \cite{HVSMR}. Each high-resolution ($\approx$0.9mm$^3$) 3D image was acquired on a 1.5T scanner (Philips Achieva), without contrast agent and using a free-breathing SSFP sequence with ECG and respiratory navigator gating. The HVSMR dataset includes blood pool and myocardium segmentations only. A trained rater manually separated all of the heart chambers and great vessels. The 20 images were categorized after visually assessing any gross morphological malformations: 4/20 severe (prior major reconstructive surgery, single ventricle, dextrocardia), 5/20 moderate (DORV, VSD, abnormal chamber shapes), and 11/20 mild (ASD, stenosis, etc.). The dataset was randomly split into 4 folds for cross-validation (15 training, 5 testing), with an equal number of mild, moderate and severe cases in each. Input images were resized to $\approx$128$\times$180$\times$144.

\vspace{3pt}
\textbf{Experiments:} In our tests, binary segmentation of each structure outperformed co-segmenting all of the heart chambers and vessels. We trained several models aimed at segmenting the aorta and left ventricle of CHD patients. \textbf{DIR} uses a single U-net to perform direct binary segmentation. \textbf{DIR-DIST} includes the Euclidean distance to the initial seed as an additional input channel. \textbf{ITER (stop)} is iterative segmentation using our RNN with automatic stopping, and \textbf{ITER (max)} simulates a user by choosing the segmentation with the best Dice coefficient after 30 iterations of our RNN. Finally, \textbf{ITER-SEG-ABL} is an ablation study with no data augmentation on the input segmentations. We tuned the architectural parameters for each experiment separately, nevertheless resulting in similar networks. All U-nets had 3 levels, 24 feature maps at the first level, and $\approx$870,000 parameters. The best network for direct segmentation of the aorta used $2 \times 2 \times 2$ max pooling (receptive field=$40^3$), while all others used $3 \times 3 \times 3$ max pooling (receptive field=$68^3$). For training, optimization using adadelta ran for 2000 epochs with a batch size of 1. For iterative segmentation, the $\mathrm{argmax}$ in eqn. (\ref{eq:inference}) is computed per voxel, by assuming that the segmentation of each voxel is conditionally independent of all other voxels given $\hst$. Segmentations were post-processed to keep only the largest island or the island containing the initial seed, for experiments in which this improves overall accuracy. Aorta segmentations were not penalized for descending aortas longer than in the gold-standard.

\begin{figure}[!t]
\vspace{-0.1in} % was -0.05in
\captionsetup{belowskip=0pt}
\centering
\includegraphics[height=1.28in]{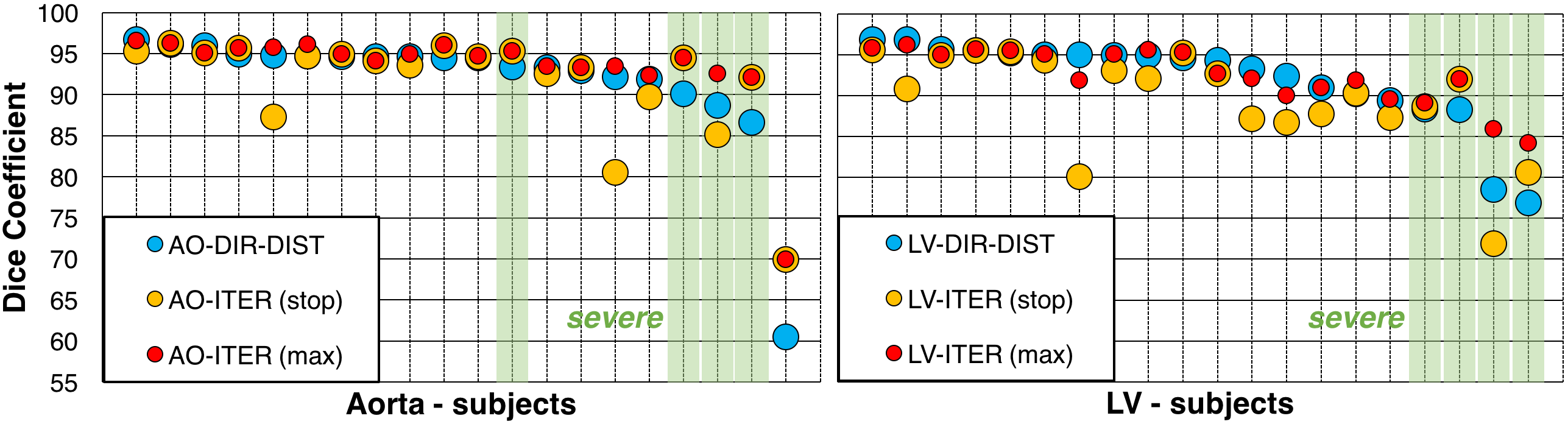} % was 1.28
\caption{Aorta (AO) and LV segmentation validation. DIR-DIST is the best direct segmentation method, but iterative segmentation generalizes better to severe subjects. Top: Dice coefficients for all methods. Bottom: Results for all 20 subjects, sorted by DIR-DIST score and with severe subjects highlighted in green.}
\label{fig:findings}
\end{figure}

\begin{figure}[!t]
\captionsetup{belowskip=0pt}
\centering
\includegraphics[width=4.6in]{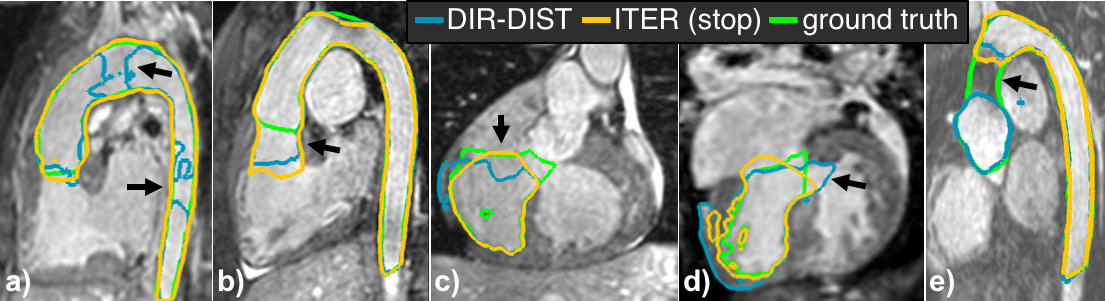} % was 4.75
\caption{Representative aorta and LV segmentations in held-out subjects with severe CHD. Arrows illustrate both the benefits and failure cases of iterative segmentation with automatic stopping, where it (a) successfully segments a difficult case, (b) stops too late, (c) correctly stops near a valve, (d) avoids growing through a septal defect, (e) cannot grow through a dark region caused by a stent.}
\label{fig:pictures}
\end{figure}

\vspace{3pt}
\textbf{Results:} Fig. \ref{fig:findings} and \ref{fig:pictures} report the results. There was no notable difference in accuracy between the mild and moderate groups. DIR-DIST was the best direct segmentation method, demonstrating the advantage of leveraging user interaction. For all methods, incorporating L-R and A-P flips in the data augmentation improved performance for severe subjects. Iterative segmentation stopped automatically after 18$\pm$3 steps for both the aorta and the LV, requiring $\approx$15 seconds. The potential benefits of our iterative segmentation approach are demonstrated by the performance of ITER (max), which shows improvement for all of the severe cases while maintaining accuracy for the others. The stopping prediction is not perfect at test time: the number of iterations separating the automatic stopping point from the best segmentation in a sequence was 0.8$\pm$1.0 iterations for the aorta and 3.0$\pm$2.5 iterations for the LV. The sole aorta containing a stent was poorly segmented by all methods (Fig. \ref{fig:pictures}e). The stent caused a strong inhomogeneity artifact that the iterative segmentation could not grow past, and the stopping criterion was never triggered.

\newcommand{\mytab}{
\begin{table}[t]
\label{tab:numbersTable}
\centering
\begin{tabularx}{4.75in}{c | c | c | c | c}
%\hline
\hline
% & \multicolumn{2}{c}{Aorta} & \multicolumn{2}{c}{LV}\\
Method & AO mild/mod. & \:\:AO severe \:\:& LV mild/mod. & \:\:LV severe\:\:\\
\hline
%ALL & 					91.5$\pm$5.2 & 75.9$\pm$8.0 & 	94.4$\pm$2.3 & 48.2$\pm$42.1 \\
DIR & 					92.5$\pm$6.5 & 81.2$\pm$16.3 & 	94.1$\pm$3.5 & 68.6$\pm$25.5 \\
DIR-DIST & 				92.3$\pm$8.6 & 89.7$\pm$2.9 & 	94.1$\pm$2.2 & 83.0$\pm$6.2 \\
\hline
ITER (stop) & 				91.5$\pm$7.0 & 91.8$\pm$4.6 & 	91.2$\pm$4.4 & 83.3$\pm$9.0 \\
ITER (max) & 				93.3$\pm$6.3 & 93.6$\pm$1.5 & 	93.7$\pm$2.3 & 87.8$\pm$3.5 \\
%\hline \hline
%ALL-NO-FLIP & 			91.8$\pm$5.1 & 58.3$\pm$8.7 & 	94.6$\pm$2.0 & 55.1$\pm$25.7 \\
%DIR-NO-FLIP & 			91.4$\pm$7.3 & 72.2$\pm$26.6 & 	94.0$\pm$3.6 & 33.7$\pm$35.1 \\
%DIR-DIST-NO-FLIP & 		92.4$\pm$7.4 & 81.8$\pm$9.4 & 	94.3$\pm$2.4 & 75.5$\pm$8.8 \\
\hline
%ITER-NO-FLIP (stop) & 		93.4$\pm$3.6 & 83.3$\pm$13.4 &  93.1$\pm$2.4 & 75.5$\pm$7.5 \\
%ITER-NO-FLIP (max) & 		93.9$\pm$2.9 & 90.7$\pm$6.6 & 	94.5$\pm$1.7 &77.0$\pm$6.9 \\
ITER-SEG-ABL (stop) & 	65.9$\pm$24.1 & 45.0$\pm$33.4 & 62.2$\pm$24.9 & 49.2$\pm$31.3 \\
ITER-SEG-ABL (max) & 	66.3$\pm$24.4 & 45.8$\pm$37.4 & 64.4$\pm$22.4 & 52.7$\pm$25.1 \\
\hline
\end{tabularx}
\end{table}}

\mytab

%\begin{figure}%
%  \centering
%  \subfloat[][]{\mytab}%
%  \qquad
%  \subfloat[][]{\mytab}
%  \caption{Here are some tables in a \texttt{figure} environment.}%
%  \label{fig:table}%
%\end{figure}

\section{Conclusions}

We presented an iterative segmentation model and its RNN implementation. We showed that for whole heart segmentation, the iterative approach was more robust to the cardiac malformations of severe CHD. Future work will investigate the potential general applicability of iterative segmentation when one is restricted to a small training dataset despite wide anatomical variability.

\vspace{3pt}
\textbf{Acknowledgements:} NSERC CGS-D, Phillips Inc., Wistron Corporation, BCH Translational Research Program and Office of Faculty Development, Harvard Catalyst, Charles H. Hood Foundation and American Heart Association.

%Future work
%\begin{itemize}
%\item rolling - output feeds into itself - whatever that stopping term is. "One alternative is to train the RNN using its own outputs, for example using XXX, in which YYY." - scheduled sampling
%\item more traditional rnn with states as input on each step (removes Markov assumption, can remember history - more complicated model??)
%\item Since we are learning  $p(\yst, \sst \vert \x, \ystmo, \sstmo = 0)$ and not $p(\yst, \sst \vert \x, \ystmo)$ the outputs $\yst$ and $\sst$ are not necessarily coupled - $\yst$ can say to keep growing while $\sst$ says to stop, or $\sst$ can say to keep growing while $\yst$ converges. Alternative is to compute the final merging $p(\yst, \sst \vert \x, \ystmo)$ inside the optimization - tried but seemed harder to optimize. Maybe using whatever that stopping term is after pretraining.
%\item other chambers
%\item Sample from h and $y_{i-1}$ instead of max value, e.g variational approaches.
%\item Integrate shape prior
%\item improve accuracy of stopping
%\item grow multiple objects at the same time
%\item additional stents
%\end{itemize}

\bibliographystyle{splncs03_sortOrder}
\bibliography{master}

\end{document}